\documentclass[11pt]{article}
\usepackage[a4paper,margin=1in]{geometry}
\usepackage[round,authoryear]{natbib}
\usepackage{amsmath,amssymb}
\usepackage{booktabs}
\usepackage{graphicx}
\usepackage{microtype}
\usepackage{xcolor}
\usepackage{url}
\usepackage{multirow}
\usepackage{tabularx}
\usepackage{float}
\usepackage{authblk}
\usepackage{hyperref}
\hypersetup{
  hidelinks,
  pdftitle={When Do Fewer Visual Tokens Accelerate Multimodal Inference? A Break-Even Study Across Decision Locations and Hardware},
  pdfauthor={Hao Dou and Ruiwen Tian}
}

\title{When Do Fewer Visual Tokens Accelerate Multimodal Inference?\\A Break-Even Study Across Decision Locations and Hardware}
\author[1]{Hao Dou}
\author[1]{Ruiwen Tian}
\affil[1]{Harbin Institute of Technology, Harbin, China}
\date{August 2026}

\begin{document}
\maketitle

\begin{abstract}
Fewer visual tokens do not guarantee lower end-to-end latency. We evaluate break-even with a reproducible protocol that accounts for decision overhead, shared work, and the operators each policy can avoid. A stage-level decomposition reconciles these components with measured end-to-end latency. In a 30-example pilot, the two tested autoregressive probes remain slower than Full despite state reuse. A lightweight post-vision predictor yields paired confidence intervals below zero on RTX~3090 and A100 and remains significant after a conservative all-pairs Holm correction. A pre-vision image-size rule also yields intervals below zero on both GPUs, although neither comparison remains significant after the same correction. Pre-vision routing has a structural opportunity unavailable to post-vision pruning: it can avoid preprocessing and vision encoding. On A100, this opportunity outweighs a nearly eightfold larger downstream token reduction by the post-vision policy. Reported quality is conditional on examples answered correctly by Full and is not benchmark accuracy.
\end{abstract}

\section{Introduction}

Multimodal large language models (MLLMs) encode images into visual-token sequences that are consumed by a language model. High-resolution inputs increase both vision-encoder work and language-model prefill, motivating a broad class of token pruning, merging, and compression methods \citep{shang2024prumerge,li2024tokenpacker,zhang2024sparsevlm,ye2024fitprune,zhang2024fastervlm}. Dynamic-resolution models such as Qwen2.5-VL expose this dependence directly because the image resolution controls the resulting token count \citep{bai2025qwen25vl}.

Efficiency claims often assume
\begin{equation}
\text{fewer visual tokens}\Rightarrow\text{less computation}\Rightarrow\text{lower latency}.
\label{eq:assumption}
\end{equation}
These implications are not equivalent. Post-vision pruning cannot recover work already spent by the vision encoder. An online gate may require an additional forward pass or decoded probe. A router can cost more than the computation it avoids. Token ratios and FLOPs therefore describe potential savings, not necessarily wall-clock behavior.

We ask when adaptive visual budgeting produces a measurable reduction in end-to-end latency, and how that outcome depends on decision overhead, decision location, and the hardware-specific balance of operators. Our study focuses on break-even rather than treating decision placement as sufficient. We do not propose a new token-importance score or claim a universally superior router.

We profile 2,600 examples from VQAv2, TextVQA, and ChartQA \citep{goyal2017vqav2,singh2019textvqa,masry2022chartqa}. The post-hoc post-vision Oracle retains 23.92\% of visual tokens on average among examples answered correctly by Full, but the smallest fixed budget satisfying the empirical validation-set preservation constraint remains at 100\%. The tested autoregressive probes do not reach latency break-even in the 30-example pilot despite state reuse. In the balanced dual-hardware timing workload, Static-0.900 has paired confidence intervals below zero and remains significant after the conservative all-pairs Holm sensitivity analysis; the Image-Size Rule also has intervals below zero but does not remain significant after that analysis. On A100, Static removes nearly eight times more visual tokens yet saves less latency than the Rule, and the Full-normalized Static--Rule latency contrast changes significantly between the two tested GPU/software environments.

The paper is an empirical study of the relationship among visual budget, avoided computation, and measured latency. It does not claim large universal acceleration, learned-router superiority, or direct conversion of the 23.92\% Oracle into system speed. The approximately 3\% Rule-level improvement is a case study for the protocol, not the principal methodological contribution. Our contributions are:
\begin{itemize}
    \item We provide a reproducible break-even protocol that records decision overhead, reusable work, avoidable operators, and a stage decomposition reconciled with measured end-to-end latency.
    \item Across two GPU/software environments, Static-0.900 yields paired latency reductions that remain significant after the conservative Holm sensitivity analysis, while the two tested autoregressive probes remain slower than Full in the pilot.
    \item The fixed-policy comparison reveals a significant Full-normalized hardware interaction and, on A100, a reversal between tokens removed and latency saved. The Rule's gains are concentrated in a high-resolution scene-text tail. Because the policies are not quality-matched, we interpret their contrast mechanistically rather than as a deployment ranking.
\end{itemize}

\section{Related Work}

Post-vision methods remove or aggregate encoded tokens through attention, learned projection, or token-importance estimates \citep{shang2024prumerge,li2024tokenpacker,zhang2024sparsevlm,ye2024fitprune,zhang2024fastervlm}. They primarily address which encoded tokens remain; our question is whether a budgeting decision reduces measured latency after its complete online cost is included. The Static Predictor is an internal low-cost representative, not a competitive reproduction of a cited pruning system, and our results do not establish the behavior of the broader post-vision literature.

Adaptive resolution changes computation before or during encoding. CARES and ResAdapt select target-model input resolution before the main encoder \citep{kimhi2026cares,liao2026resadapt}, while AdaptVision conditionally acquires detailed crops from a compressed input \citep{lin2026adaptvision}. Qwen2.5-VL maps variable-resolution images to variable-length sequences through its \texttt{max\_pixels} interface \citep{bai2025qwen25vl}, which we use for controlled pre-vision routing. These approaches differ in decision location, additional policy work, and the operators they can avoid. We compare tested representatives at pre-vision, post-vision/pre-generation, and autoregressive-online locations, but location is neither randomized nor isolated: each representative is a different implementable policy. We therefore interpret location through operator avoidability and measured accounting, not as a controlled causal factor.

\section{Budget, Avoided Computation, and Latency}
\label{sec:motivation}

\subsection{Post-vision safe budgets}

For image $I$, question $q$, and retained-token ratio $r$ in
\begin{equation}
\mathcal{R}=\{0.05,0.10,0.15,0.25,0.50,0.75,1.00\},
\end{equation}
we define the monotone-safe post-vision budget of a Full-correct example as
\begin{equation}
r^*_{\mathrm{safe}}=\min\left\{r\in\mathcal{R}:S(y_{r'},y^*)>0\quad\forall r'\geq r\right\},
\label{eq:safe}
\end{equation}
where $S$ is the dataset-specific answer score. Thus, profile ``correctness'' means positive VQA soft score for VQAv2/TextVQA and exact relaxed-numeric correctness for ChartQA; no unspecified threshold $\tau$ is used. The monotonic requirement prevents an accidental low-budget success from being labeled safe when a larger budget fails.

The complete profile contains 2,600 examples and 18,200 runs. Full inference answers 1,962 examples correctly. Their mean safe budget is 23.92\%; 41.74\% are safe at 5\%, and only 4.13\% require 100\%. The non-monotonic rate is 14.02\%. Mean safe budgets differ substantially by dataset: 16.50\% for VQAv2, 18.01\% for TextVQA, and 41.07\% for ChartQA. Under a 98.5\% monotone-safe validation constraint, however, the selected fixed budget is 100\%.

The 23.92\% Oracle is defined after visual encoding and does not quantify a pre-vision resolution Oracle. It measures semantic redundancy in encoded tokens. Pre-vision routing exploits only the narrower subset of work that can be avoided by reducing the input resolution.

\subsection{A general empirical break-even accounting framework}

For any adaptive policy, we define
\begin{equation}
T_{\mathrm{adaptive}}=T_{\mathrm{shared}}+T_{\mathrm{decision\text{-}only}}+T_{\mathrm{execution\text{-}only}},
\label{eq:adaptive-time}
\end{equation}
where $T_{\mathrm{shared}}$ is computation executed once that serves both the decision and subsequent inference; $T_{\mathrm{decision\text{-}only}}$ contains feature aggregation, policy inference, rule computation, routing, and token selection used only to choose and enact a budget; and $T_{\mathrm{execution\text{-}only}}$ contains the remaining executed preprocessing, vision-encoder, and combined LLM prefill/decode work. These categories are mutually exclusive. For Static, image preprocessing and vision encoding up to the reusable visual features are shared: the processor outputs and visual features are each computed once and reused. Static's additional image statistics and feature aggregation are included in decision-only work. For the pre-vision Image-Size Rule, shared computation is approximately zero. The empirical break-even test is
\begin{equation}
T_{\mathrm{adaptive}}<T_{\mathrm{full}}.
\label{eq:break_even}
\end{equation}
This is an explicit accounting framework and reusable measurement test, not a theoretical guarantee: decision location limits which operators are avoidable, while overhead determines whether those potential savings remain at the end-to-end boundary. We distinguish three evidence levels. \emph{Observed break-even} means a negative measured mean latency difference relative to Full and does not by itself imply statistical support. \emph{Paired statistical support} means that the sample-level paired confidence interval for method minus Full excludes zero on the negative side. \emph{Family-wise sensitivity support} means that the comparison remains significant under the conservative all-pairs Holm sensitivity analysis.

\begin{table*}[t]
\caption{Instantiation of the accounting boundary by policy.}
\label{tab:accounting-policies}
\centering
\scriptsize
\begin{tabularx}{\textwidth}{lXXX}
\toprule
Policy & Decision cost includes & Potentially avoidable operators & Shared computation \\
\midrule
Confidence Gate & Probe prefill/decode, gate logic & Remaining LLM processing & Visual features, prefix, KV \\
Counterfactual Gate & Two probe paths, comparison & Remaining LLM processing & Visual features, prefix, KV \\
Static Predictor & Features, MLP, token selection & Combined LLM prefill/decode & Image preprocessing and vision encoding \\
Image-Size Rule & Geometry-only routing & Preprocess, vision, LLM processing & Approximately zero \\
\bottomrule
\end{tabularx}
\end{table*}

We evaluate a Confidence Gate and a Counterfactual Gate. Both begin from an initial reduced context; the latter adds a compact region around a highly relevant unseen token and measures whether the short generated prefix changes. The binary target is \emph{budget sufficiency}: positive if the initial retained-token ratio is at least the monotone-safe post-vision budget $r^*_{\mathrm{safe}}$, and negative otherwise. The Counterfactual improvement over Confidence is not significant (AUROC difference $0.0060$, 95\% CI $[-0.0194,0.0343]$), while the second probe adds 136.0~ms on average.

Table~\ref{tab:probe} reports an exploratory RTX~3090 pilot after visual-feature, prefix, and KV reuse. Its 30 examples comprise ten from each dataset and are used to illustrate decision cost rather than to estimate dual-hardware performance. Both tested autoregressive probes fail observed break-even because their measured mean latency exceeds Full, whereas the non-autoregressive Static Predictor clears observed break-even in this pilot. The result motivates separating decision cost from decision location.

\begin{table}[t]
\caption{Exploratory integrated online-probe benchmark on RTX~3090: 30 examples (ten per dataset), three repeats, and five Full warm-ups. Quality is descriptive for this pilot; latency differences are paired after within-sample averaging.}
\label{tab:probe}
\centering
\small
\begin{tabular}{lrrrr}
\toprule
Method & Mean latency (ms) & Relative latency & Delta vs. Full (ms) & 95\% CI (ms) \\
\midrule
Full & 401.3 & 1.000 & -- & -- \\
Static Predictor & 386.5 & 0.963 & $-14.9$ & $[-19.6,-10.5]$ \\
Confidence Gate & 472.5 & 1.177 & $+71.1$ & $[+50.1,+91.3]$ \\
Counterfactual Gate & 605.4 & 1.508 & $+204.0$ & $[+180.3,+229.3]$ \\
\bottomrule
\end{tabular}
\end{table}

The Static Predictor maps already-computed visual features and lightweight question statistics to one of seven retained-token ratios without decoding a probe. It cannot avoid vision encoding, but its cost can remain below the combined LLM prefill/decode work it removes. We therefore include it in the subsequent dual-hardware evaluation in Section~\ref{sec:placement-results}.

\section{Pre-Vision Routing}
\label{sec:method}

\subsection{A family of pre-vision policies}

Pre-Vision Routing is a family of policies that selects an input-resolution tier before invoking the vision encoder. In the main analysis, we evaluate fixed tiers, an Image-Size Rule, and a Question + Metadata Router. A validation-selected conditional trigger is retained as a negative result in the appendix. None requires an image embedding or generated probe.

The main pre-vision policy is the validation-locked Image-Size Rule (hereafter, the Rule):
\begin{equation}
\widehat{N}_{\mathrm{full}}(I)=\left\lceil\frac{w}{28}\right\rceil\left\lceil\frac{h}{28}\right\rceil,
\end{equation}
\begin{equation}
\pi_{\mathrm{size}}(I)=
\begin{cases}
602112,&\widehat{N}_{\mathrm{full}}(I)\geq1097,\\
802816,&\text{otherwise.}
\end{cases}
\label{eq:size_rule}
\end{equation}
The threshold is selected on 157 Full-correct validation examples subject to an empirical validation-set relative three-task macro constraint of $\geq0.985$. This is an observed validation criterion, not a population non-inferiority guarantee. The Rule uses no dataset identity, vision encoder, or LLM call. Under the final timing protocol, its complete online routing path costs approximately 0.012~ms on RTX~3090 and 0.011~ms on A100; this interval includes geometry evaluation and tier dispatch.

The geometric estimate is available before processor execution and is used only for routing; its relationship to processor tokenization is audited in Appendix~\ref{sec:processor-audit}. All reported token and latency results use processor/model records.

The Question + Metadata Router uses a frozen question embedding, lexical statistics, and image geometry, but not dataset identity. We report all three predeclared training seeds and use the model to test whether learning provides a stable conditional-quality advantage over the simpler Rule. It is not the primary deployment policy.

A validation-selected conditional trigger provides no measurable improvement over the always-applied rule; its definition and complete results are reported in Appendix~\ref{sec:conditional-appendix}.

\section{Experimental Protocol}
\label{sec:setup}

We use Qwen2.5-VL-3B-Instruct in bfloat16 with batch size one, greedy decoding, and at most 24 output tokens. VQAv2 and TextVQA use soft accuracy; ChartQA uses relaxed numerical accuracy.

The fixed evaluation comprises 403 Full-correct examples (160 VQAv2, 133 TextVQA, and 110 ChartQA), evaluated under four pre-vision resolution tiers; frozen Static policies use post-vision retained-token tiers. Here, Full-correct means a positive dataset-specific score as defined in Eq.~\ref{eq:safe}, not necessarily a score of one. Conditioning therefore allows VQAv2 and TextVQA Full means below one, whereas binary relaxed correctness fixes the ChartQA Full mean at one. The evaluation was frozen before the final closure analyses but had been inspected during earlier development, so we treat it as a fixed conditional evaluation rather than an untouched test set. Its scores measure conditional failure avoidance among requests answered correctly by Full, not benchmark accuracy or behavior on Full-error requests. The predeclared three-task macro remains primary, with the non-ceiling macro auxiliary.

Repeated latency uses 120 examples, stratified into 12 dataset-by-within-dataset-Full-token-quartile cells with ten examples per cell. The primary latency estimand is therefore the equally weighted mean over these 12 predeclared cells. It characterizes the balanced timing workload, not the natural dataset mixture or a deployment request distribution. Each method receives 20 warm-ups and five randomized repeats per sample. We average repeats within sample, use 10,000 paired-bootstrap resamples for confidence intervals, and use paired sign-flip tests for latency hypotheses. The bootstrap resamples requests from this empirical balanced workload without reweighting to a deployment distribution; the sign-flip test treats paired differences as sign-exchangeable under its null. The request is the analysis unit, so the intervals reflect workload heterogeneity rather than uncertainty across machines or deployment instances. Fully online latency includes preprocessing, online policy work, routing, token selection, vision, and LLM execution; only common image-file loading is excluded. Complete sampling, software, hardware, and checkpoint details are reported in Appendix~\ref{sec:reproducibility}.

\section{Results}
\label{sec:results}

\subsection{No stable conditional-quality advantage from learning is detected}

Table~\ref{tab:quality} reports how well each method preserves answer scores on the fixed conditional evaluation. Fixed-602112 fails the empirical validation-set criterion because TextVQA drops to 0.9073, while the Rule has the same aggregate conditional quality as Fixed-802816 despite routing 19 high-resolution examples to the lower 602112 tier (Appendix~\ref{tab:fixed-tiers}). None of the three predeclared seeds provides consistent paired-bootstrap or McNemar evidence of a conditional-quality advantage over the Rule. Most routing disagreements do not change the answer score; complete seed results are in Appendix~\ref{sec:router-seeds}.

\begin{table*}[t]
\caption{Answer score on the fixed 403-example Full-correct conditional evaluation. Three-task macro is primary; non-ceiling macro is auxiliary. The Router is the mean of three seeds. The empirical 0.985 validation-set constraint does not transfer to Static-0.900 on this evaluation.}
\label{tab:quality}
\centering
\scriptsize
\setlength{\tabcolsep}{3.5pt}
\begin{tabular}{lrrrrrr}
\toprule
Method & VQAv2 & TextVQA & ChartQA & 3-task macro & NC macro & Rel. Full \\
\midrule
Full & 0.9688 & 0.9825 & 1.0000 & 0.9837 & 0.9756 & 1.0000 \\
Static-0.900 & 0.9563 & 0.9549 & 0.9909 & 0.9673 & 0.9556 & 0.9833 \\
Image-Size Rule & 0.9688 & 0.9674 & 1.0000 & 0.9787 & 0.9681 & 0.9949 \\
Question + Metadata Router & 0.9688 & $0.9708\pm0.0072$ & 1.0000 & $0.9798\pm0.0024$ & $0.9698\pm0.0036$ & $0.9960\pm0.0025$ \\
\bottomrule
\end{tabular}
\end{table*}

ChartQA contributes task coverage and can expose degradation, but its Full score is fixed at a conditioning-induced ceiling. We therefore also report the auxiliary non-ceiling macro
\begin{equation}
Q_{\mathrm{macro,NC}}=\frac{Q_{\mathrm{VQAv2}}+Q_{\mathrm{TextVQA}}}{2}.
\end{equation}
This diagnostic does not replace the predeclared three-task macro.

A validation-threshold sensitivity analysis produces no distinct routing or quality operating point; complete calibration results are reported in Appendix~\ref{sec:static-calibration}.

\subsection{Tested low-cost policies and break-even}
\label{sec:placement-results}

On the balanced timing workload, Static-0.900 reduces paired latency by 27.12~ms on RTX~3090 and 4.02~ms on A100. Both confidence intervals exclude zero, and both comparisons remain significant after the conservative all-pairs Holm correction. The Rule reduces latency by 13.99~ms on RTX~3090 and 8.92~ms on A100. Its confidence intervals also exclude zero, although neither comparison remains significant after the same correction (Table~\ref{tab:latency}; Appendix~\ref{sec:pairwise-holm}). For sample $i$ and hardware $h$, let $d_{ih}=(T^{\mathrm{Static\mbox{-}0.900}}_{ih}-T^{\mathrm{Rule}}_{ih})/T^{\mathrm{Full}}_{ih}$. The interaction estimand is
\begin{equation}
\Delta_{\mathrm{int}}=\frac{1}{n}\sum_i(d_{i,\mathrm{RTX}}-d_{i,\mathrm{A100}}).
\label{eq:hardware-interaction}
\end{equation}
The same 120 sample identifiers are paired across hardware; bootstrap resampling draws these paired identifiers, and the sign-flip test randomizes the paired cross-hardware contrasts. Negative values indicate a larger Static advantage relative to the Rule on RTX~3090. The Full-normalized interaction estimate is $-0.04364$, 95\% CI $[-0.05152,-0.03627]$, paired sign-flip $p<10^{-5}$. This is a fixed-policy interaction across the two tested GPU/software environments, not a causal effect of decision location. Because the policies are not quality-matched, we interpret it mechanistically rather than as a deployment ranking.

\begin{table*}[t]
\caption{Fully online latency for the balanced 12-cell timing workload, with five repeats per sample. $\Delta$ is method minus Full; negative values indicate lower latency. Quality is relative three-task macro on the separate fixed 403-example conditional evaluation.}
\label{tab:latency}
\centering
\scriptsize
\setlength{\tabcolsep}{4.5pt}
\begin{tabular}{llrrrrrr}
\toprule
Hardware & Method & Loc. & Mean (ms) & Rel. lat. & Delta (ms) & 95\% CI (ms) & Rel. score \\
\midrule
\multirow{3}{*}{RTX 3090} & Full & -- & 482.87 & 1.0000 & 0.00 & $[0,0]$ & 1.0000 \\
& Static-0.900 & Post & 455.75 & 0.9438 & $-27.12$ & $[-32.62,-21.96]$ & 0.9833 \\
& Image-Size Rule & Pre & 468.88 & 0.9710 & $-13.99$ & $[-26.45,-3.96]$ & 0.9949 \\
\midrule
\multirow{3}{*}{A100 PCIe} & Full & -- & 320.86 & 1.0000 & 0.00 & $[0,0]$ & 1.0000 \\
& Static-0.900 & Post & 316.84 & 0.9875 & $-4.02$ & $[-6.43,-1.82]$ & 0.9833 \\
& Image-Size Rule & Pre & 311.94 & 0.9722 & $-8.92$ & $[-16.78,-2.71]$ & 0.9949 \\
\bottomrule
\end{tabular}
\end{table*}

On the locked timing subset, Static-0.900 retains 423.5 visual tokens versus 619.0 for Full and spends 3.22~ms (RTX~3090) or 2.84~ms (A100) on decision work. The Rule lowers mean processor tokens to 594.0 and can avoid preprocessing and vision work.

\paragraph{Removing more tokens need not save more time.} On A100, Static-0.900 removes 195.5 visual tokens on average, nearly eight times the 25.0 removed by the Rule, yet saves only 4.02~ms compared with 8.92~ms. Both quantities refer to visual tokens presented to the language model: retained encoded tokens for Static and processor-produced tokens for the resolution rule. Pre-vision routing has a structural opportunity unavailable to post-vision pruning because it can avoid preprocessing and vision encoding. Here, that opportunity outweighs the much larger downstream token reduction by Static, which acts after those expensive operators have executed.

The stage decomposition in Table~\ref{tab:break-even-components} reconciles with the measured end-to-end latency difference. The larger combined LLM prefill/decode reduction accounts for most of Static's RTX saving, while the pre-vision Rule derives a larger share from measured vision reduction on A100. The A100 Static residual is non-negligible relative to its small net saving. The decomposition therefore supports an operator-mix interpretation, but not precise sub-millisecond causal attribution to individual stages. The pattern is consistent with different operator proportions across the two tested environments.

Equation~\ref{eq:adaptive-time} partitions absolute adaptive latency into mutually exclusive categories. Table~\ref{tab:break-even-components} instead decomposes the measured Full-minus-method latency difference by execution stage. Shared work is executed once and therefore is not reported as a separate saving term.

Static and Rule are not quality-matched: their relative Full macros are 0.9833 and 0.9949. The 12-comparison Holm family is a conservative all-pairs sensitivity analysis; complete results appear in Appendix~\ref{sec:pairwise-holm}. The quality--latency operating-point figure is reported in Appendix~\ref{sec:supp-quality-latency}.

\begin{table*}[t]
\caption{Full-minus-method accounting reconciliation from the final per-run records. Decision-only contains policy and budget-enactment work, including token selection. For Static, image preprocessing and vision encoding are shared and executed once; for Rule, shared work is approximately zero. Shared work is not a separate saving term. Net and stage reductions are Full minus method, so positive values indicate savings. LLM-stage reduction is the recorded combined LLM prefill/decode interval because v3 did not persist a valid split. Residual is computed as the end-to-end net reduction minus the classified stage terms and decision cost; it makes the components sum to the measured difference but is not an independent validation of stage attribution. $^*$ Static preprocessing and vision differences are measured variation rather than structurally avoidable work, because Static acts only after full visual encoding.}
\label{tab:break-even-components}
\centering
\scriptsize
\resizebox{\textwidth}{!}{%
\begin{tabular}{llrrrrrrr}
\toprule
Hardware & Method & Decision-only & Preproc. red. & Vision red. & LLM-stage red. & Residual red. & Net red. & Closed \\
\midrule
RTX 3090 & Static-0.900 & 3.225 & $-0.207^*$ & 0.961$^*$ & 28.603 & 0.989 & 27.122 & yes \\
& Image-Size Rule & 0.012 & 0.843 & 12.020 & 1.138 & 0.005 & 13.994 & yes \\
\midrule
A100 & Static-0.900 & 2.839 & $-0.149^*$ & 0.400$^*$ & 5.778 & 0.828 & 4.019 & yes \\
& Image-Size Rule & 0.011 & 0.698 & 7.462 & 0.771 & $-0.001$ & 8.920 & yes \\
\bottomrule
\end{tabular}}
\end{table*}

\subsection{Where does the speedup come from?}
\label{sec:concentration}

The six timing examples routed to 602112 account for 98.01\% of signed RTX saving; after removing them, the remaining 114 examples save only 0.29~ms on average. The corresponding A100 remainder is $-0.01$~ms. The Rule's mean benefit is therefore a workload-mixture effect and is approximately absent outside these six routed requests. Bootstrap and leave-one-out analyses show that the concentration is not driven by any single request; complete statistics are reported in Appendix~\ref{sec:concentration-audit}.

\begin{figure*}[t]
\centering
\includegraphics[width=.35\textwidth]{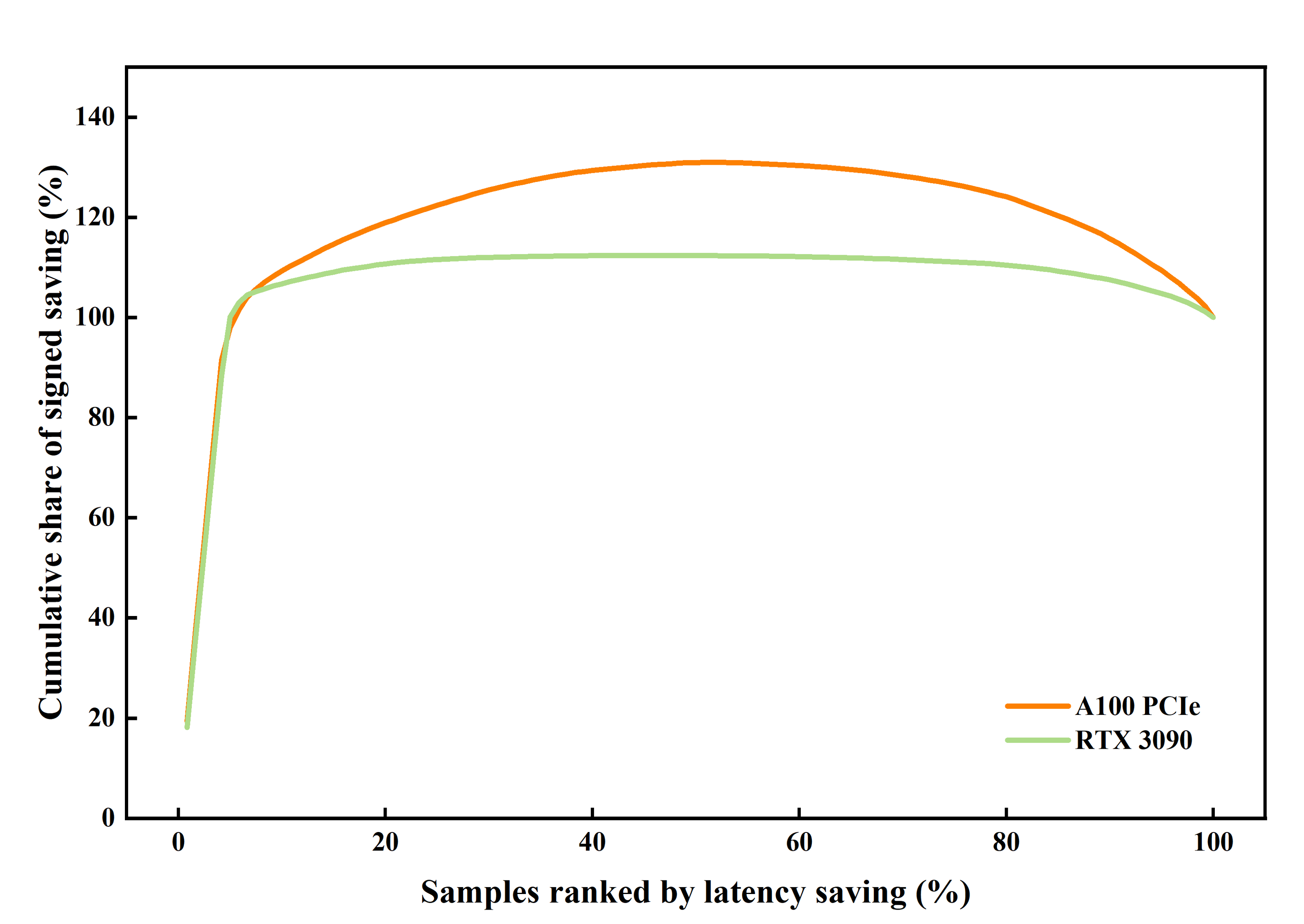}\hfill
\includegraphics[width=.63\textwidth]{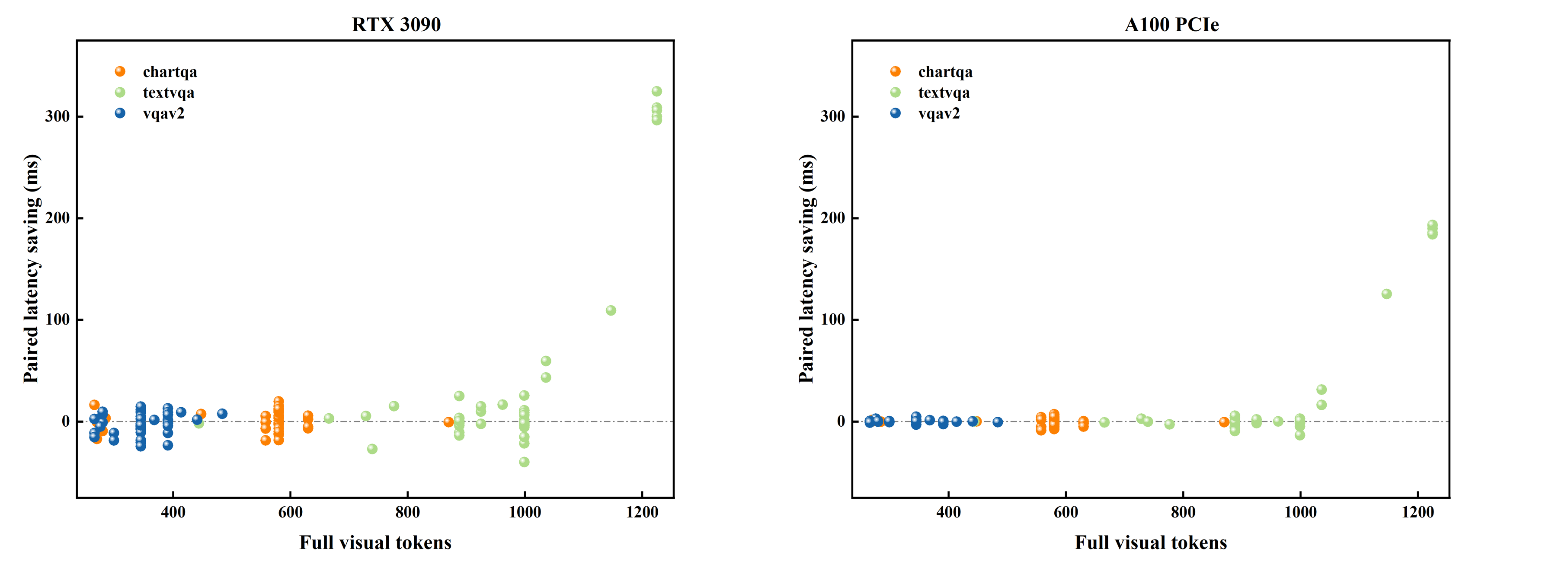}
\caption{Image-Size Rule latency-saving concentration. Left: cumulative signed saving after ranking samples by paired saving. Curves may exceed 100\% because negative-saving samples reduce the final total. Right: Full visual tokens versus paired Rule saving.}
\label{fig:concentration}
\end{figure*}

All 19 fixed-evaluation examples assigned to 602112 are from TextVQA. Their mean visual-token count drops from 1,208.4 to 734.6. In the six corresponding timing examples, mean RTX~3090 saving is 270.2~ms. Yet their mean answer score falls from 0.9649 to 0.8596. High-resolution scene-text images offer the largest observed latency opportunity while also carrying the greatest quality risk.

\section{Discussion}

\subsection{Measurement and operator boundaries}
Break-even provides a useful measurement framework: it maps each policy to a complete timing boundary, decision overhead, reusable work, avoidable operators, and an accounting residual reconciled with end-to-end latency. The tested probes fail in the pilot because decoding overhead exceeds avoided work. Static acts after vision encoding, whereas the Rule can avoid processor and encoder work; hence removing more tokens can save less time. The Full-normalized fixed-policy interaction further shows that latency contrasts depend on the hardware/software operator mix, although two environments do not identify a unique causal mechanism.

\subsection{Quality and workload boundaries}
Static and the Rule are not quality-matched, so their latency levels are not deployment rankings. Rule savings concentrate in a small high-resolution TextVQA tail that also carries the principal scene-text quality loss. Full-correct conditioning creates a ChartQA ceiling, Conditional triggering provides no measurable gain, and the learned Router shows no stable conditional-quality advantage over the Rule. The findings apply most directly to workloads containing enough high-resolution inputs for pre-vision routing to avoid substantial computation.

\section{Limitations}

The study covers one model family, three datasets, batch size one, two GPU/software environments, and repeated latency on a balanced 120-example conditional workload. The 403-example Full-correct evaluation measures conditional failure avoidance rather than benchmark accuracy and does not characterize Full-error requests. Static is an internal post-vision representative, not a published pruning baseline, and the results do not generalize automatically to production batching or other serving stacks. Decision location is not isolated experimentally because each location is represented by a different policy.

The Rule's savings concentrate in 19 high-resolution TextVQA examples in the fixed evaluation and six examples in the timing workload; their prevalence in deployment traffic is unknown. BF16 greedy answers vary slightly across GPUs, so quality is anchored to the fixed 403-example evaluation and latency is reported separately by environment. Finally, the post-hoc post-vision Oracle and pre-vision routing control different quantities; the $100-23.92=76.08$ percentage-point Oracle gap is not a pre-vision latency opportunity.

\section{Conclusion}

The safe post-vision token budget varies substantially across examples: the post-hoc Oracle retains 23.92\% of visual tokens on average, while the smallest fixed budget satisfying the empirical validation-set constraint remains at 100\%. The two tested autoregressive probes remain slower than Full in the 30-example pilot despite state reuse. On the balanced dual-hardware workload, Static and the Rule both have negative paired latency estimates with confidence intervals excluding zero, but only Static remains significant after the conservative all-pairs Holm sensitivity analysis. Pre-vision routing can avoid preprocessing and vision encoding, a structural opportunity unavailable to post-vision pruning. On A100, that opportunity outweighs a nearly eightfold larger downstream token reduction by Static. Token counts cannot replace operator-level wall-clock measurement.

\subsubsection*{Broader Impact Statement}
Reducing unnecessary visual computation may lower inference cost and energy use. Resolution reduction can also remove small text or other safety-relevant detail. Deployments should measure subgroup quality and should not infer safety from aggregate token or latency reductions.

\subsubsection*{Ethics Declaration}
The study uses publicly available research datasets and pretrained models. It collects no new human-subject data and does not infer sensitive personal attributes. Source dataset licenses and usage conditions remain applicable.

\subsubsection*{Data and Code Availability}
Code for profiling, routing, paired latency evaluation, and aggregation will be released with the final manuscript. Experiments use public VQAv2, TextVQA, and ChartQA datasets and public Qwen2.5-VL checkpoints. Derived profiles will be released where source licenses permit.

\subsubsection*{Conflict of Interest}
The authors declare no known competing interests.

\bibliography{references}
\bibliographystyle{plainnat}

\appendix

\section{Reproducibility and Environment Details}
\label{sec:reproducibility}

The 120-example subset contains ten examples per dataset and within-dataset Full-token quartile. Its unweighted request mean is therefore the equally weighted mean of the 12 cells. Quartiles are computed over all 403 examples separately within each dataset, with rank-based tie breaking and sampling seed 20260720. Because tied token counts are split by rank, the four cells need not have disjoint observed token ranges. Processor-token ranges are ChartQA: Q1 256--558, Q2 580, Q3 580, Q4 580--870; TextVQA: Q1 333--888, Q2 888--999, Q3 999, Q4 999--1,225; and VQAv2: Q1 256--345, Q2 345, Q3 345--391, Q4 391--529.

Image-identity hashing was used to construct the development and fixed-evaluation splits. The 403 fixed-evaluation examples have no dataset--sample identifier overlap with the 443 training and 157 validation examples used for router development. No filtering used router outputs, latency measurements, or final-policy behavior.

The final runs used an NVIDIA A100 PCIe 40GB and, after instance migration, an NVIDIA GeForce RTX~3090 with driver 570.195.03. The bundle preserved the slow Qwen processor, SDPA implementation, Python 3.11.10, PyTorch 2.4.0+cu121, Transformers 4.51.3, CUDA 12.1, cuDNN 9.1.0, generation parameters, scoring code, sample manifest, and identical Static checkpoint (SHA-256 prefix \texttt{4a14b263}). All thresholds were selected on validation data. Method order was randomized within each sample and repeat with seed 20260720. CUDA synchronization brackets the recorded GPU stages. Records were flushed per sample and method; repeats were averaged within sample before paired analysis, making the request rather than the individual repeat the inferential unit.

\section{Safe-Budget Distribution}

\begin{table}[h]
\caption{Distribution of monotone-safe post-vision budgets among 1,962 Full-correct examples.}
\centering
\small
\begin{tabular}{rrrrrrrr}
\toprule
Budget & 5\% & 10\% & 15\% & 25\% & 50\% & 75\% & 100\% \\
\midrule
Count & 819 & 249 & 167 & 249 & 250 & 147 & 81 \\
Rate & .4174 & .1269 & .0851 & .1269 & .1274 & .0749 & .0413 \\
\bottomrule
\end{tabular}
\end{table}

\section{Fixed Resolution Tier Results}

\begin{table}[h]
\caption{All four fixed pre-vision resolution tiers evaluated on the fixed 403-example conditional evaluation. Relative Full is the three-task macro divided by Full's three-task macro.}
\label{tab:fixed-tiers}
\centering
\small
\begin{tabular}{lrrrrr}
\toprule
Fixed tier & VQAv2 & TextVQA & ChartQA & 3-task macro & Relative Full \\
\midrule
Fixed-401408 & 0.9688 & 0.8571 & 0.9545 & 0.9268 & 0.9421 \\
Fixed-602112 & 0.9688 & 0.9073 & 1.0000 & 0.9587 & 0.9745 \\
Fixed-802816 & 0.9688 & 0.9674 & 1.0000 & 0.9787 & 0.9949 \\
Fixed-1003520 (Full) & 0.9688 & 0.9825 & 1.0000 & 0.9837 & 1.0000 \\
\bottomrule
\end{tabular}
\end{table}

\section{Processor and Geometric Proxy Audit}
\label{sec:processor-audit}

The factor 28 in $\widehat{N}_{\mathrm{full}}$ reflects Qwen2.5-VL patching and spatial merging. The processor subsequently applies aspect-ratio-preserving resizing, minimum/maximum pixel constraints, and divisibility rules. Accordingly, the geometric proxy is used only before processor execution for routing. On the fixed evaluation, its Pearson correlation with the recorded Full processor-token count is 0.994, while their ranges are 112--1,369 and 256--1,225. All formal token and latency results use processor/model records rather than the proxy.

\section{Online-Probe Implementation Detail}

The integrated implementation encodes visual features once per method invocation. If a gate accepts, generation continues from the probe prefix and KV state unless an end-of-sequence token has already been produced. If it rejects, Full-budget generation reuses the visual representation. A continuation smoke test matches one-shot generation. The Counterfactual Gate adds a compact neighborhood around a validation-selected unseen region and compares token, logit, confidence, and hidden-state changes. These checks reduce avoidable recomputation but do not reverse the negative latency result.

\subsection{Static Predictor}

The Static Predictor is a post-vision multilayer perceptron with hidden widths 256 and 128. Its inputs concatenate a frozen question embedding and lexical statistics with the already-computed global image embedding, image statistics, and question--image cross features. It predicts seven cumulative indicators, one for whether each retained-token tier is sufficient under the monotone-safe label in Eq.~\ref{eq:safe}. Decision thresholds are selected on validation data, and inference chooses the smallest tier whose cumulative probability clears its threshold. Thus, ``static'' means that no autoregressive probe is decoded; it does not mean pre-vision.

Budget prediction and budget execution are separate operations. After selecting a ratio, the implementation retains $\lceil rN\rceil$ encoded visual tokens. Seventy-five percent of the retained capacity is assigned, where possible, to tokens with the highest cosine similarity to the mean question embedding; the remainder is allocated to approximately uniform centers on the merged visual grid to preserve spatial coverage. Duplicate indices are removed, retained indices are restored to spatial order, and all text and special tokens remain unchanged. The same hybrid selector is used in the seven-tier profile and Static execution. Processor outputs and visual features are each computed once and reused by the policy and language model. Feature aggregation, cosine scoring, coverage selection, top-$k$, indexing, and budget enactment are included in decision-only time. Because obtaining the image embedding has already paid the preprocessing and vision-encoder costs, Static can avoid only downstream language-model work. Table~\ref{tab:latency} shows that this saving exceeds its measured decision cost in both tested environments.

\section{Static Threshold Calibration}
\label{sec:static-calibration}

Table~\ref{tab:static-calibration} audits the two frozen Static thresholds. Thresholds were selected exclusively on the 157-example validation split. The fixed 403-example evaluation was used only after locking for this threshold comparison. The search was restricted to thresholds in $[0.900,1.000)$, so that the calibrated point could not be more aggressive than the prespecified Static-0.900 reference. Thresholds below 0.900 were excluded by design. The search evaluated 342 candidates from a fixed grid and every predictor-probability decision point in this interval; 328 satisfied the empirical validation-set relative three-task macro criterion of $\geq0.985$.

The minimum validation mean retained-token count was 417.306, attained by exactly two candidates, 0.900000000 and 0.900431693, which produced identical validation routes and quality. The recorded selection code first minimizes mean retained tokens and then selects the larger probability threshold on an exact tie; this uniquely yields Static-Val985 at 0.900431693. The two thresholds differ on three fixed-evaluation routes, increasing mean retained tokens from 416.007 to 418.238 without changing any answer score. Static-Val985 is therefore a validation-constrained deterministic tie-break point, not an independent quality operating point.

\begin{table*}[t]
\caption{Static threshold calibration audit. Changed routes are measured relative to Static-0.900. Validation and evaluation token means are computed on the 157-example validation split and the fixed 403-example conditional evaluation. The 423.5 and 426.1 token values reported in the latency section instead refer to the balanced 120-example timing workload. Evaluation values were observed only after threshold locking.}
\label{tab:static-calibration}
\centering
\scriptsize
\setlength{\tabcolsep}{3.2pt}
\begin{tabularx}{\textwidth}{lrrrrrrX}
\toprule
Point & Threshold & Val. rel. macro & Val. tokens & Eval. tokens & Val. changed & Eval. changed & Role \\
\midrule
Static-0.900 & 0.900000000 & 0.9885 & 417.306 & 416.007 & 0 & 0 & Prespecified aggressive reference \\
Static-Val985 & 0.900431693 & 0.9885 & 417.306 & 418.238 & 0 & 3 & Validation-constrained deterministic tie-break point \\
\bottomrule
\end{tabularx}
\end{table*}

\begin{table*}[t]
\caption{Calibration-sensitivity stage decomposition for Static-Val985. Definitions and signs match Table~\ref{tab:break-even-components}; token selection is included only in decision-only work, while image preprocessing and vision encoding are shared. LLM-stage reduction is the combined LLM prefill/decode reduction. $^*$ Static preprocessing and vision differences are measured variation rather than structurally avoidable work, because the policy acts only after full visual encoding.}
\label{tab:static-val985-decomposition}
\centering
\scriptsize
\begin{tabular}{lrrrrrrr}
\toprule
Hardware & Decision-only & Preproc. red. & Vision red. & LLM-stage red. & Residual red. & Net red. & Closed \\
\midrule
RTX 3090 & 3.053 & 0.155$^*$ & 0.965$^*$ & 28.442 & 0.984 & 27.493 & yes \\
A100 & 2.584 & 0.038$^*$ & 0.497$^*$ & 5.739 & 0.835 & 4.525 & yes \\
\bottomrule
\end{tabular}
\end{table*}

\begin{table*}[t]
\caption{Static-Val985 calibration-sensitivity latency and quality. $\Delta$ is method minus Full; quality is relative three-task macro on the fixed conditional evaluation.}
\label{tab:static-val985-latency}
\centering
\small
\begin{tabular}{lrrrrr}
\toprule
Hardware & Mean (ms) & Relative latency & Delta (ms) & 95\% CI (ms) & Relative Full \\
\midrule
RTX 3090 & 455.38 & 0.9431 & $-27.49$ & $[-32.91,-22.44]$ & 0.9833 \\
A100 PCIe & 316.34 & 0.9859 & $-4.53$ & $[-6.85,-2.32]$ & 0.9833 \\
\bottomrule
\end{tabular}
\end{table*}

Static-Val985 is a deterministic validation tie-break point. It does not create a distinct validation or fixed-evaluation quality operating point. As a calibration-sensitivity check, the Full-normalized Static-Val985-minus-Rule cross-hardware interaction is $-0.04261$, with 95\% CI $[-0.04964,-0.03578]$ and paired sign-flip $p<10^{-5}$. This is not a parallel main interaction result.

\section{Conditional Pre-Vision Trigger}
\label{sec:conditional-appendix}

Conditional triggering is a validation-selected extension retained as a negative result. Its prespecified three-branch policy is
\begin{equation}
\pi_{\mathrm{cond}}(I)=
\begin{cases}
1003520, & \widehat{N}_{\mathrm{full}}(I)<682,\\
802816, & 682\leq\widehat{N}_{\mathrm{full}}(I)<1097,\\
602112, & \widehat{N}_{\mathrm{full}}(I)\geq1097.
\end{cases}
\label{eq:conditional}
\end{equation}
Both thresholds were locked on the 157-example Full-correct validation split by minimizing latency subject to validation-level relative three-task macro preservation $\geq0.985$. The non-Full branches trigger on 35.73\% of the fixed 403-example evaluation and 36.67\% of the timing subset. Conditional and Always Rule have identical observed conditional quality. Conditional minus Always is $+0.59$~ms on RTX~3090 (95\% CI $[-1.48,2.69]$, Holm $p=1.0$) and $+0.14$~ms on A100 ($[-0.47,0.75]$, Holm $p=1.0$). The final conclusion is that the trigger provides no measurable improvement; Always Rule remains the representative pre-vision policy.

\section{Question + Metadata Router Seed-Level Tests}
\label{sec:router-seeds}

All three seeds were predeclared and are reported without test-based selection. Table~\ref{tab:router-seeds} compares each seed directly with Image-Size Rule. The paired-bootstrap difference is learned minus rule in absolute macro quality. McNemar counts use $n_{10}$ for rule-correct/learned-wrong and $n_{01}$ for rule-wrong/learned-correct. The intervals and exact tests do not establish non-inferiority or equivalence; they support only the narrower statement that no stable significant learned-router advantage was detected.

\begin{table*}[t]
\caption{Seed-level Question + Metadata Router tests on the fixed 403-example evaluation set. Bootstrap uses 10,000 sample-level resamples; McNemar $p$ is exact and two-sided.}
\label{tab:router-seeds}
\centering
\small
\begin{tabular}{rrrrrrr}
\toprule
Seed & Relative Full & Macro difference & 95\% CI & $n_{10}$ & $n_{01}$ & McNemar $p$ \\
\midrule
20260720 & 0.9975 & 0.00251 & $[0.00000,0.00752]$ & 0 & 1 & 1.000 \\
20260721 & 0.9975 & 0.00251 & $[0.00000,0.00752]$ & 0 & 1 & 1.000 \\
20260722 & 0.9932 & $-0.00167$ & $[-0.00919,0.00585]$ & 2 & 1 & 1.000 \\
\bottomrule
\end{tabular}
\end{table*}

\section{Locked Latency Pairwise Tests}
\label{sec:pairwise-holm}

Table~\ref{tab:pairwise-holm} reports a conservative all-pairs sensitivity analysis among Full, Static-0.900, Static-Val985, and the Rule on both GPUs. Two-sided paired sign-flip tests use 200,000 randomizations, and Holm correction covers all six pairs on each GPU, 12 comparisons in total. This broad family is a sensitivity analysis rather than the only possible multiplicity definition. Cross-hardware interactions test different estimands and are reported separately in Section~\ref{sec:placement-results}.

\begin{table}[H]
\caption{All prespecified latency pairwise comparisons. $\Delta$ is first minus second after within-sample averaging.}
\label{tab:pairwise-holm}
\centering
\scriptsize
\begin{tabular}{lllrrrr}
\toprule
Hardware & First & Second & Delta (ms) & 95\% CI (ms) & Raw $p$ & Holm $p$ \\
\midrule
RTX 3090 & Full & Static-0.900 & 27.12 & $[21.96,32.62]$ & $<10^{-5}$ & 0.00006 \\
& Full & Static-Val985 & 27.49 & $[22.44,32.91]$ & $<10^{-5}$ & 0.00006 \\
& Full & Rule & 13.99 & $[3.96,26.45]$ & 0.01393 & 0.11140 \\
& Static-0.900 & Static-Val985 & 0.37 & $[-0.70,1.45]$ & 0.50361 & 0.70366 \\
& Static-0.900 & Rule & $-13.13$ & $[-23.51,-0.66]$ & 0.02356 & 0.13323 \\
& Static-Val985 & Rule & $-13.50$ & $[-24.18,-1.05]$ & 0.02221 & 0.13323 \\
\midrule
A100 & Full & Static-0.900 & 4.02 & $[1.82,6.43]$ & 0.00024 & 0.00216 \\
& Full & Static-Val985 & 4.53 & $[2.32,6.85]$ & 0.00002 & 0.00015 \\
& Full & Rule & 8.92 & $[2.71,16.78]$ & 0.01585 & 0.11140 \\
& Static-0.900 & Static-Val985 & 0.51 & $[-0.19,1.30]$ & 0.18850 & 0.70366 \\
& Static-0.900 & Rule & 4.90 & $[-1.18,12.20]$ & 0.17591 & 0.70366 \\
& Static-Val985 & Rule & 4.39 & $[-1.77,11.87]$ & 0.22789 & 0.70366 \\
\bottomrule
\end{tabular}
\end{table}

\section{Saving Concentration Audit}
\label{sec:concentration-audit}

On RTX~3090, the top 10\% and 20\% of samples contribute 109.28\% and 118.93\% of signed Rule saving, exceeding 100\% because slower samples offset positive savings; TextVQA contributes 102.99\%. On A100, the six 602112 samples, top 10\%, top 20\%, and TextVQA contribute 100.07\%, 106.74\%, 110.71\%, and 101.28\%. On RTX~3090, 48.33\% of samples are slower under the Rule, while 21.67\%, 8.33\%, and 5.83\% save more than 10, 20, and 50~ms. The A100 values are 54.17\%, 6.67\%, 5.83\%, and 5.00\%.

The six 602112 timing samples save 1,645.8~ms in total on RTX~3090. A 10,000-resample bootstrap restricted to them gives $[1,239.4,1,882.7]$~ms; leave-one-out totals range from 1,320.9 to 1,536.6~ms, and the largest sample contributes 19.74\% of the total. These analyses show that the concentrated tail effect is not created by one request alone.

\section{Dataset and Effective-Tier Saving Breakdown}
\label{sec:dataset-tier-breakdown}

\begin{figure}[H]
\centering
\includegraphics[width=.98\columnwidth]{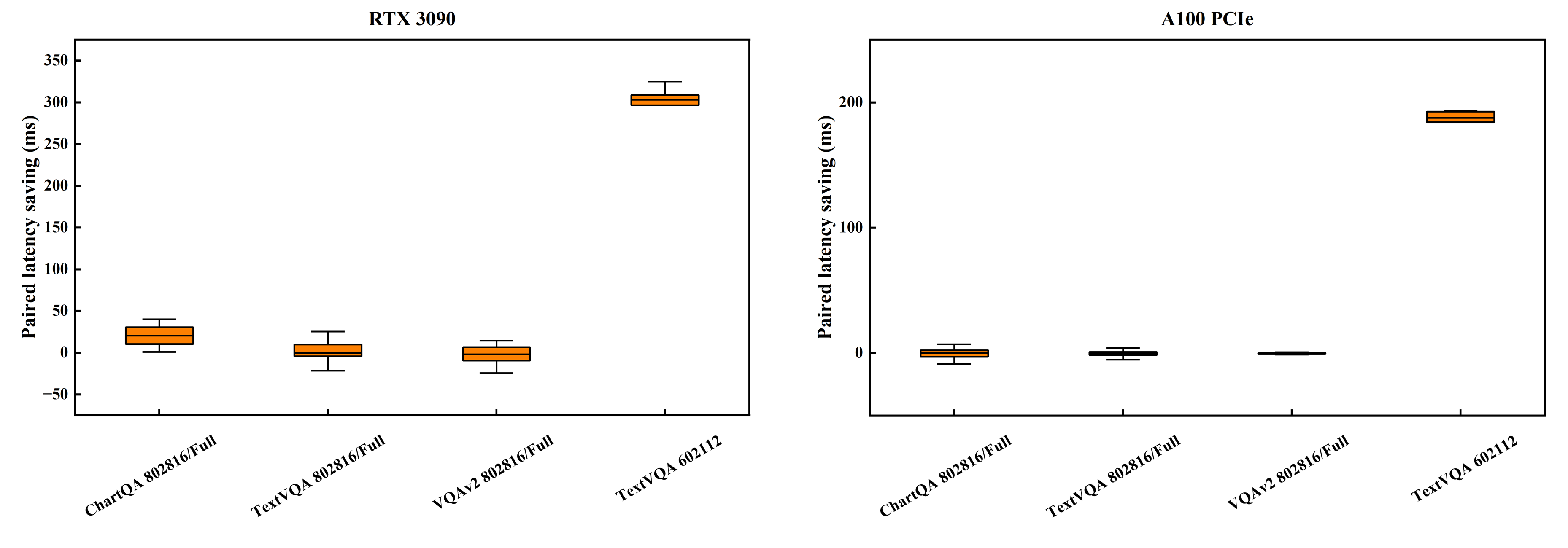}
\caption{Paired latency saving grouped by dataset and effective tier. The 602112 group contains high-resolution TextVQA examples and dominates net saving.}
\label{fig:groups}
\end{figure}

\section{High-Resolution Tail Audit}
\label{sec:tail-audit}

Table~\ref{tab:tail19} lists every fixed-evaluation example routed to 602112. A dash in the latency column means that the example was not selected into the 120-example repeated-latency workload; no profile-time latency is substituted. This prevents the 19-example conditional score audit from being conflated with the six-example repeated timing result.

\begin{table}[H]
\caption{Per-example audit of the 19 high-resolution TextVQA inputs routed to 602112. Tokens are processor-reported counts; score is Full $\rightarrow$ routed; saving is paired Image-Size Rule saving on RTX~3090.}
\label{tab:tail19}
\centering
\scriptsize
\begin{tabular}{rrrr@{\qquad}rrrr}
\toprule
ID & Tokens & Score & Saving (ms) & ID & Tokens & Score & Saving (ms) \\
\midrule
16552 & 1225$\to$729 & 1.000$\to$1.000 & -- & 1776 & 1225$\to$729 & 1.000$\to$1.000 & -- \\
18953 & 1225$\to$729 & 1.000$\to$1.000 & -- & 20620 & 1225$\to$729 & .667$\to$.667 & -- \\
20662 & 1147$\to$750 & 1.000$\to$1.000 & -- & 24249 & 1225$\to$729 & .667$\to$.667 & -- \\
26970 & 1225$\to$729 & 1.000$\to$1.000 & -- & 27336 & 1221$\to$754 & 1.000$\to$0.000 & -- \\
29034 & 1225$\to$729 & 1.000$\to$1.000 & -- & 30565 & 1225$\to$729 & 1.000$\to$1.000 & 308.95 \\
31060 & 1225$\to$729 & 1.000$\to$1.000 & -- & 3151 & 1225$\to$729 & 1.000$\to$1.000 & 306.10 \\
32733 & 1073$\to$744 & 1.000$\to$1.000 & -- & 34348 & 1225$\to$729 & 1.000$\to$1.000 & 324.87 \\
35156 & 1221$\to$754 & 1.000$\to$1.000 & -- & 36086 & 1225$\to$729 & 1.000$\to$1.000 & -- \\
36850 & 1147$\to$750 & 1.000$\to$0.000 & 109.25 & 4116 & 1225$\to$729 & 1.000$\to$1.000 & 300.09 \\
5067 & 1225$\to$729 & 1.000$\to$1.000 & 296.56 & & & & \\
\bottomrule
\end{tabular}
\end{table}

\section{Supplementary Quality--Latency Figure}
\label{sec:supp-quality-latency}

\begin{figure}[H]
\centering
\includegraphics[width=.92\columnwidth]{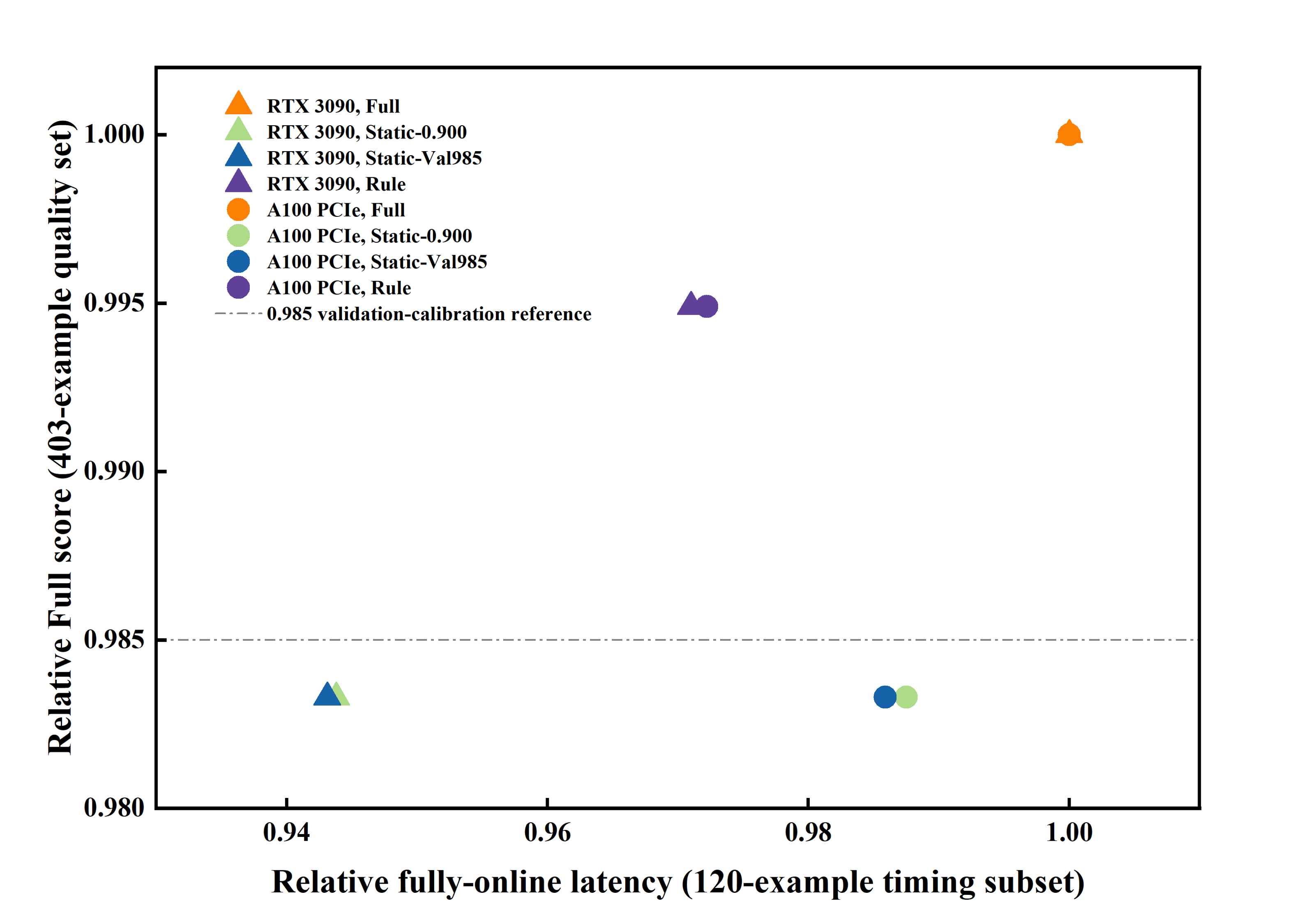}
\caption{Observed operating points. Quality is measured on the fixed 403-example Full-correct conditional evaluation, whereas latency is measured on the balanced 120-example timing workload. Quality and latency are therefore separate prespecified endpoints, not a same-sample Pareto frontier. Both Static points fall below the 0.985 validation-calibration reference after transfer.}
\label{fig:quality-latency}
\end{figure}

\end{document}